\documentclass{article}

    \usepackage[preprint]{neurips_2025}

\usepackage[utf8]{inputenc} 
\usepackage[T1]{fontenc}    
\usepackage{hyperref}       
\usepackage{url}            
\usepackage{booktabs}       
\usepackage{graphicx}       
\usepackage{pifont}
\newcommand{\cmark}{\ding{51}}
\usepackage{amsmath}        
\usepackage{amsfonts}       
\usepackage{nicefrac}       
\usepackage{microtype}      
\usepackage{xcolor}         
\usepackage{makecell}
\usepackage{listings}
\usepackage{multirow}

\title{SVQA-R1: Reinforcing Spatial Reasoning in MLLMs via View-Consistent Reward Optimization}

\author{%
  Peiyao Wang \qquad \qquad Haibin Ling\\
  Department of Computer Science, Stony Brook University\\
    \texttt{\{peiyaowang, hling\}@cs.stonybrook.edu} \\
}

\begin{document}

\maketitle

\begin{abstract}
Spatial reasoning remains a critical yet underdeveloped capability in existing vision-language models (VLMs), especially for Spatial Visual Question Answering (Spatial VQA) tasks that require understanding relative positions, distances, and object configurations. Inspired by the R1 paradigm introduced in DeepSeek-R1, which enhances reasoning in language models through rule-based reinforcement learning (RL), we propose SVQA-R1, the first framework to extend R1-style training to spatial VQA. In particular, we introduce Spatial-GRPO, a novel group-wise RL strategy that constructs view-consistent rewards by perturbing spatial relations between objects, e.g., mirror flipping, thereby encouraging the model to develop a consistent and grounded understanding of space. Our model, SVQA-R1, not only achieves dramatically improved accuracy on spatial VQA benchmarks but also exhibits interpretable reasoning paths even without using supervised fine-tuning (SFT) data.
Extensive experiments and visualization demonstrate the effectiveness of SVQA-R1 across multiple spatial reasoning benchmarks. 
\end{abstract}

\section{Introduction}

Although open-source vision language models (VLM), such as LLaVA~\cite{liu2023visual} and Qwen2.5-VL~\cite{bai2025qwen2}, have achieved impressive results in standard VQA tasks, they often fail to handle spatial reasoning well~\cite{chen2024spatialvlm}. These limitations stem from several issues. First, such models exhibit a strong local visual bias, tending to focus on salient individual objects rather than modeling spatial relations between them. For example, when asking ``Is the cup to the left of the laptop'', the model may successfully identify the cup and the laptop separately but fail to reason about their relative positions. This is partially because most existing training data prioritize object classification over spatial logic, leading to an underdeveloped spatial reasoning capacity. As a result, models may rely heavily on object categories and make arbitrary spatial guesses. Second, spatial VQA tasks often require fine-grained localization, which general VLMs struggle with, especially in cluttered scenes or when small objects are involved. Finally, these models typically lack an explicit reasoning chain, jumping directly from visual input to answer without intermediate thinking steps. This makes them prone to failure in multi-step spatial understanding, such as interpreting ``above the fridge and on the table'' as a layered spatial relation. These limitations collectively make general VLMs ill-suited for complex spatial understanding tasks.

Recent works~\cite{chen2024spatialvlm, ma2024spatialpin} have begun addressing these limitations by introducing spatially-targeted supervision. For example, SpatialVLM~\cite{chen2024spatialvlm} proposes a synthetic data generation pipeline to efficiently construct large-scale spatial VQA samples, allowing supervised training that improves performance in spatial reasoning tasks. Building on this, SpatialPIN~\cite{ma2024spatialpin} incorporates 3D priors to directly enhance object relation modeling without modifying the underlying VLM architecture, which yields further gains. However, both approaches rely heavily on supervised fine-tuning (SFT), which may lead to convergence to static output patterns and limit reasoning flexibility. To overcome these constraints, we draw inspiration from the R1 paradigm and explore a reinforcement learning-based approach that encourages dynamic self-correcting spatial reasoning behavior through rule-based feedback.

In the R1 paradigm proposed by DeepSeek-R1~\cite{guo2025deepseek}, simple rule-based reinforcement learning (RL) has been shown to significantly enhance the reasoning capabilities of large language models (LLMs) without the need for manually annotated data thought processes. This approach offers an efficient and scalable pathway for training complex reasoning skills. Building upon this paradigm, we systematically extend the R1 framework to spatial reasoning tasks and propose the \textbf{SVQA-R1} framework. Specifically, we tailor the R1-style reinforcement learning to the characteristics of vision-language models by introducing a \textbf{Spatial-GRPO} mechanism. In this setup, we construct a rule-based reward to verify the format correctness and a semantic-aware reward to handle the verification of different answer types, \textit{e.g.}, bounding box, caption, distance, single or multiple choice, \textit{etc}. Furthermore, we design a view-consistent reward that encourages the policy model to yield the correct answer for both the original VQA and the spatially augmented one at the same time, in case the model guesses only one of them right by chance. Thus, our proposed Spatial-GRPO can offer the model a view-consistent spatial reasoning capacity.

In the experiment, we observe that SVQA-R1 can improve the accuracy over the SFT-based baseline by more than \textbf{30\%} on the Q-Spatial++ benchmark~\cite{liao2024reasoning}, while also exhibits an interpretable reasoning path. In addition, it dramatically outperforms the closed-source models Gemini-1.5-Flash~\cite{gemini-1.5-flash} and it is competitive with the GPT-4o~\cite{islam2024gpt}. Also, it outperforms open-source models InternVL-2.5~\cite{chen2024expanding} and Qwen2.5VL~\cite{bai2025qwen2} by a large margin. We expect that such RL-based research exploration can advance the spatial reasoning in MLLMs.

Our contributions are as follows: \textbf{1)} We explore rule-based and semantic-based RL in MLLMs for the spatial reasoning task, empowered with an interpretable reasoning path via LLM prompting; \textbf{2)} we design a novel Spatial-GRPO mechanism which prompts the policy model to learn a view-consistent action space; and \textbf{3)} our proposed SVQA-R1 approach achieves a very promising result of \textbf{58\%} on Q-Spatial++, outperforming other open-source models by a large margin.

\section{Related Work}
\vspace{-1.9mm}\paragraph{Spatial Reasoning with VLMs}

Spatial reasoning, crucial for interpreting spatial relationships in visual scenes, remains a significant challenge for vision-language models (VLMs) in tasks like spatial visual question answering (Spatial VQA). Early VLMs, such as CLIP~\cite{radford2021learning}, excel in 2D visual tasks but struggle with 3D spatial relationships and compositional reasoning~\cite{wang2024picture,patel2024tripletclip,doveh2023dense}. Models like MDETR\cite{kamath2021mdetr} improve cross-modal alignment with spatial attention, yet quantitative spatial tasks (e.g., distance estimation) remain difficult\cite{liao2024reasoning}. Recent advancements address these limitations: SpatialVLM~\cite{chen2024spatialvlm} leverages a 2-billion-example 3D spatial dataset, enhancing VLMs’ qualitative and quantitative Spatial VQA capabilities, supporting applications like chain-of-thought (CoT) reasoning~\cite{wei2022chain}. SpatialPIN~\cite{ma2024spatialpin} introduces zero-shot 3D reasoning via 3D priors, improving Spatial VQA and robotics tasks like trajectory planning. MM-Spatial~\cite{daxberger2025mm} utilizes large-scale 3D scene data with the Cubify Anything VQA (CA-VQA) dataset, enabling robust 3D spatial understanding, including spatial relationship prediction and metric estimation, with state-of-the-art performance on 3D benchmarks. Datasets like GQA~\cite{hudson2019gqa} and Visual Spatial Reasoning (VSR)~\cite{liu2023visual} highlight VLM weaknesses in multi-hop spatial queries. While neuro-symbolic methods\cite{yi2018neural,vedantam2019probabilistic} and RL-guided attention\cite{salter2021attention,hong2021reinforced} decompose spatial tasks or optimize decisions, explicit reasoning via CoT and adaptive learning through RL are underexplored for complex Spatial VQA. Our work aims to bridge this gap, enhancing VLM generalization and compositional reasoning in diverse spatial scenarios.

\paragraph{Chain-of-Thought and R1-style Reinforcement Learning In Vision-Language Models}

Recent research has increasingly focused on enhancing the reasoning capabilities of vision-language models through structured prompting and reinforcement learning. Chain-of-Thought (CoT) prompting~\cite{wei2022chain}, initially effective in language models, has been extended to vision-language models to improve multi-step visual reasoning, grounding, and spatial inference~\cite{zhao2023cotvlm,wang2022raven}. 
Complementary to prompting-based approaches, R1-style reinforcement learning originally proposed in DeepSeek-R1~\cite{guo2025deepseek} has been adapted to multimodal domains to promote verifiable, rule-guided learning. Models such as Visual-RFT~\cite{wu2024visualrft} and VLM-R1~\cite{zhou2024vlm} apply Group Relative Policy Optimization (GRPO)~\cite{shao2024deepseekmath} with task-specific rewards (e.g., correctness of spatial relations or grounding accuracy), achieving superior generalization over supervised fine-tuning. Video-R1~\cite{xu2024videor1} further extends this framework to temporal reasoning by integrating T-GRPO and constructing fine-grained video-reasoning datasets. Recent extensions, such as TinyLLaVA-Video-R1~\cite{li2024tinyllava} and VideoChat-R1~\cite{zhang2024videochat}, demonstrate that R1-style training enhances spatial and temporal understanding even in smaller-scale models. In addition, Reason-RFT~\cite{wu2023reasonrft} explores reasoning-aware RL using chain-of-thought supervision combined with GRPO.

Together, these CoT-based prompting and R1-style RL methods reflect a shift towards more explicit and interpretable reasoning in multimodal systems, enabling models to generalize better across complex spatial, temporal, and causal reasoning tasks.

\section{Method}

We begin by presenting our approach to generating mirror-consistent QA pairs in Section~\ref{QA}, and then introduce the mixed reward strategy in Section~\ref{reward} and the overall Spatial-GRPO training procedure in Section~\ref{GRPO}.

\subsection{Mirror-Consistent Reasoning via QA Adaptation}
\label{QA}
To encourage the policy model to learn a view-consistent action space, we can use Nerf~\cite{mildenhall2021nerf} or Gaussian Splatting~\cite{kerbl20233d} to synthesize novel-view images for the existing data, or collect the multi-view data~\cite{yu2023mvimgnet} from scratch. Due to the imperfect synthetics and expensive data collection costs, we instead use mirror flipping to augment the spatial relation for the single-view image datasets. Such a low-cost operation allows us to easily scale our approach to any existing dataset at a large scale.

To enable mirror-consistent spatial reasoning, we construct horizontally flipped image samples paired with semantically aligned question-answer (QA) pairs. A key challenge is that simple horizontal flipping changes the spatial configuration (e.g., ``left'' becomes ``right''), requiring corresponding adjustments to the associated QA content to ensure consistency.

\begin{figure*}[!t]
  \centering
   \includegraphics[width=\linewidth]{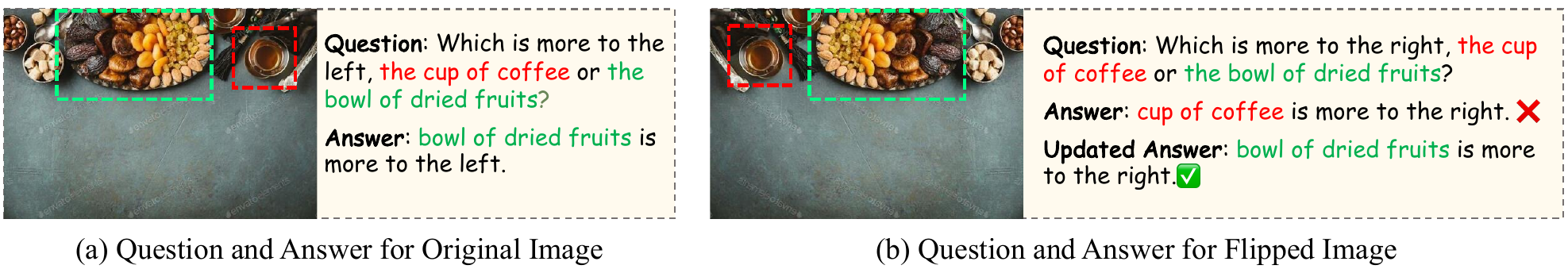}
   \caption{(a) The original image, question, and answer of a sample. (b) The flipped image, question, and the answers before and after verification enhancement.}
   \label{fig:flip}
\end{figure*}

To address this, we employ GPT-4o~\cite{islam2024gpt} to automatically generate revised QA pairs for each flipped image. We provide the original image's question and answer, and prompt the model to produce a logically correct version suitable for the flipped view. However, we observe that a naïve prompting strategy often leads to shallow edits. GPT-4o tends to mechanically swap ``left'' with ``right'' without properly understanding the full relational semantics of the scene. This results in numerous logically inconsistent or even contradictory QA pairs, illustrated in Figure~\ref{fig:flip} (b) as the answer of the flipped image.

To improve reliability, we enhance the prompt by explicitly instructing GPT-4o to verify the consistency of the revised QA with the mirrored image. The prompt is shown in Table~\ref{tab:prompt}, and the red highlight is the enhanced content. This includes guidance to reason about relative positions and object identities before finalizing the output. With this enhanced prompting strategy, most flipped QA pairs are logically correct and spatially coherent, even in multi-object scenes involving complex spatial arrangements, illustrated in Figure~\ref{fig:flip} (b) as the updated answer of the flipped image.

This QA adaptation process enables us to generate high-quality mirrored-view data, providing the necessary foundation for implementing Spatial-GRPO. It ensures that both original and flipped samples are semantically aligned, making joint reward computation meaningful and accurate during training.
\begin{table}[tbp]
\renewcommand{\arraystretch}{1.3} 
\centering
\label{tab:prompt}
\footnotesize 
\begin{tabular}{p{0.9\linewidth}}
\hline
You are a spatial reasoning assistant. You are given a question and its corresponding answer based on an image. Now assume that the image has been horizontally flipped. Your task is to rewrite the question and answer so that they remain logically correct for the flipped image. Write them \textbf{as if the flipped image was the original}, and \textbf{do not mention the flip} in your output. \textcolor{red}{Also, verify the correctness of the left/right spatial relationship in the original answer. If the rewritten answer is inconsistent with the horizontal flip (i.e., the object that was on the left is still on the left), you must fix it. If you find an error, correct the object-direction mapping accordingly. } Original question: \{\textcolor{blue}{question}\}. Original answer: \{\textcolor{blue}{answer}\}. Return your output as a \textbf{valid JSON object}, and nothing else. \\
\hline
\end{tabular}
\caption{Instruction prompt for the flipping data. The placeholders \{\textcolor{blue}{question}\} and \{\textcolor{blue}{answer}\} will be replaced with specific content. Text in \textcolor{red}{red} highlights the verification and correction instructions.}
\end{table}

\subsection{Mixed Reward Design}
\label{reward}
\begin{figure*}[tb]
  \centering
   \includegraphics[width=\linewidth]{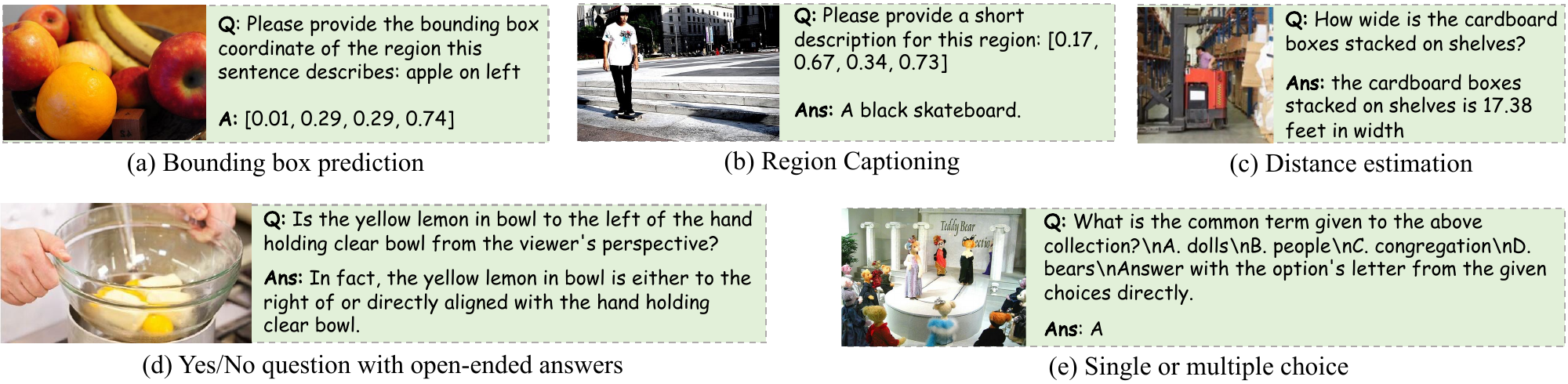}
   \caption{A diverse set of open-ended spatial question-answer types.}
   \label{fig:type}
\end{figure*}

Unlike structured tasks such as classification or numerical prediction, spatial reasoning in vision-language models (VLMs) often involves open-ended question-answering with diverse and linguistically rich output. As illustrated in Figure~\ref{fig:type}, model responses may vary in form, ranging from short directional terms to complete relational descriptions. This inherent variability presents challenges for rule-based reward functions, which rely on exact string matching or rigid templates. Applying such methods directly risks penalizing semantically correct answers that deviate from expected wording, thereby limiting the expressiveness and performance potential of VLMs.

To accommodate this flexibility, we adopt a semantic-based reward that evaluates the alignment between the model’s output and the ground truth based on meaning rather than surface form. This approach allows the model to generate natural and fluent responses while still being guided by accurate spatial understanding. In addition, motivated by findings from recent work~\cite{chen2024spatialvlm} that highlight the benefits of Chain-of-Thought (CoT) reasoning in improving multistep spatial inference, we incorporate a format reward to further encourage structured and interpretable outputs. Together, these components support a more flexible and faithful reward mechanism tailored to the open-ended nature of spatial reasoning tasks.

\paragraph{Format Reward $r^f$.} We include a binary signal that assigns 1 if the output follows the required structural format (e.g., \texttt{<think>...</think> <answer>...</answer>}), and 0 otherwise.

\paragraph{Semantic-aware Reward $r^s$.} To capture semantic similarity beyond surface-level token overlap, we use a Sentence-BERT-based reward. Specifically, we compute sentence embeddings using a pre-trained Sentence-BERT~\cite{reimers2019sentence} model (e.g., all-MiniLM-L6-v2) and measure the cosine similarity between the predicted and reference answers. This embedding-based reward enables the model to recognize semantically equivalent expressions that lexical metrics such as BLEU fail to identify, for example, assigning a high reward to predictions like "sofa" when the ground truth is "couch". In addition, Sentence-BERT-based reward can measure the numerical difference when handling those answer types, such as distances and bounding boxes. Such a reward unification can stabilize the training, as we observe. 

\paragraph{Final Reward.} The total reward is defined as a weighted sum of all components:
\begin{equation}
    r = \lambda_1 \cdot r^f + \lambda_2 \cdot r^s 
\end{equation}

where $\lambda_1, \lambda_2 $ are hyperparameters.
This composite reward encourages the model to generate answers that are structurally valid, lexically and semantically correct, and numerically plausible.

\subsection{Spatial Group Relative Policy Optimization (Spatial-GRPO)}
\label{GRPO}
While GRPO has demonstrated promising results in general reasoning tasks, it lacks explicit reward signals that are specifically designed to guide spatial reasoning. This makes it inadequate for training Vision-Language Models (VLMs) to understand and reason about spatial relationships such as ``left of'', ``above'', or ``closer to''. To address this limitation, we introduce Spatial Group Relative Policy Optimization (Spatial-GRPO), a joint reinforcement learning framework that explicitly targets spatial understanding.


Our approach is grounded in two key observations about human spatial cognition: (1) quantitative measurements such as distance between objects remain invariant under horizontal mirroring, and (2) qualitative relationships like ``in front of'' or ``next to'' are preserved, while directional expressions such as ``left'' and ``right'' require symmetric adjustment. We aim to replicate this perceptual stability, encouraging the model to maintain semantically accurate spatial reasoning for both views.


Specifically, given an original image and question pair $(I, Q)$, and the horizental fliped imge and update queation is tuple $(\hat{I}, \hat{Q})$. They will obatian grouped sample outputs $\{o_i\}^G_{i=1}$ and  $\{\hat{o}_i\}^G_{i=1}$ from the old policy model $\pi_{\text{old}}$. 
Then we compute the rewards  $r_i, \hat{r}_i$ for each sample in the original and flipped groups independently, based on their respective question-answer alignments. To promote consistent and accurate spatial reasoning across both the original and mirrored views, we aim to minimize the discrepancy between the two reward distributions. Specifically, we compare the aggregated rewards $\mathtt{Avg}(\{r_i\}_{i=1}^G)$ and $\mathtt{Avg}(\{\hat{r}_i\}_{i=1}^G)$ from the original and flipped groups and penalize the group that achieves a substantially higher score. This design reflects our assumption that a truly spatially grounded model should perform similarly across both views, and that large discrepancies indicate a lack of spatial consistency.

To operationalize this idea, we define a consistency-aware joint reward that encourages both high performance and inter-view agreement. Specifically, we use the semantic reward difference between the original and flipped views to quantify the discrepancy in model responses. This choice reflects our focus on semantic alignment, as it is independent of surface-level formatting variations in the generated outputs. We define the semantic difference as:
\begin{align}
    \Delta = \mathtt{Avg}(\{r_i^s\}_{i=1}^G) - \mathtt{Avg}(\{\hat{r}^s_i\}_{i=1}^G)
\end{align}
The the modified reward for the semantic rewards as follows:
\begin{align}
    r_i^s &= r_i^s -\eta|\Delta|, \quad \text{if} \quad r_i^s > \delta \quad \text{and} \quad \Delta >= 0  \\
    \hat{r}_i^s &= \hat{r}_i^s - \eta|\Delta|, \quad \text{if} \quad \hat{r}_i^s >\delta \quad \text{and} \quad \Delta < 0
\end{align}


 The final Spatial-GRPO maximizes the following objective:
 \begin{align}
&\mathcal{J}_{\text{Spatial-GRPO}}(\theta) = \mathbb{E}_{q,\hat{q}, \{o_i,\hat{o}_i\} \sim \pi_{\text{old}}} \left[
    \frac{1}{2G} \sum_{i=1}^{G} (R_i +\hat{R}_i) \right] - \beta D_{\text{KL}}(\pi_\theta \parallel \pi_{\text{ref}})
     \\
     R_i & = \frac{1}{|o_i|} \sum_{t=1}^{|o_i|} 
    \min \left( \alpha_t(\theta) A_{i}, \, \text{clip}(\alpha_t(\theta), 1 - \epsilon, 1 + \epsilon) A_{i} \right) , \alpha_t(\theta) = \frac{\pi_\theta(o_{i,t} \mid q, o_{i,<t})}{\pi_{\text{old}}(o_{i,t} \mid q, o_{i,<t})}
 \end{align}

where $A_{i} = \frac{r_i-\text{mean}(\{r_i\}^G_{i=1})}{\text{std}(\{r_i\}^G_{i=1})}$ is the estimated advantage, $\eta$ is a scaling coefficient, $\beta$ is the KL penalty coefficient, and $\pi_\theta$, $\pi_{\text{old}}$, $\pi_{\text{ref}}$ denote the current, old, and reference policy models, respectively. $\epsilon$ is the threshold for clipping. $q$ is the input pair of the image and the question. The definition of $\hat{R}_i$ is similar to $R_i$.

\section{Experiment}
\subsection{Experimental Settings}

We adopt Qwen2.5-VL-3B~\cite{bai2025qwen2} as our base model and use its pre-trained weights. The maximum generation length is set to 2048 tokens.
For the reward computation, we set the weighting coefficients $\lambda_1$ and $\lambda_2$ to 0.5, balancing between the semantic and format rewards. When updating the semantic reward, we apply a threshold $\delta = 0.5$, and set the final reward scaling factor $\eta = 1$.
During GRPO training, we generate 8 candidate answers per input as a group. We set the gradient accumulation steps to 2 and the per-device batch size to  8. Training is conducted on  8 NVIDIA A6000 GPUs.

\subsection{Datasets}
\paragraph{Training Dataset.}
For training, we adopt the \textbf{Vqasynth\_Spacellava}~\cite{VQASynth} dataset, which contains more than 28,000 multi-turn dialogues designed for visual question answering (VQA). This dataset includes a rich mixture of spatial VQA and general VQA samples.
All training samples are automatically generated using the VQASynth~\cite{VQASynth} framework, guided by the spatial prompting techniques introduced in SpatialVLM~\cite{chen2024spatialvlm}. We split the multi-turn dialogues into multiple single-turn data, resulting in xxx training samples in total.


\paragraph{Test Dataset.} For numerical task evaluation, we adopt the \textbf{Q-Spatial++}~\cite{liao2024reasoning} dataset. Q-Spatial++ is a subset of the Q-Spatial-Bench dataset, designed to evaluate quantitative spatial reasoning in large vision-language models. It comprises 87 freshly captured images and 101 human expert-annotated questions, focusing exclusively on horizontal distances between objects in real-world scenes. To ensure high precision, physical measurements of object distances were taken during image capture, providing accurate ground-truth answers.

In addition, to verify the generalizability of our method, we construct two additional test sets. One is based on \textbf{Vqasynth\_Spacellava}~\cite{VQASynth} and the other is based on \textbf{OpenSpaces}~\cite{VQASynth}. Both are generated using the VQASynth pipeline.
The OpenSpaces dataset is built by synthesizing spatial VQA samples from the first 30K rows of the localized narratives split of the Cauldron dataset using VQASynth, while Vqasynth\_Spacellava samples are collected directly via multi-turn chat interactions. For each source, we collect approximately 1,000 multi-turn dialogues and convert them into single-turn QA pairs in the format of (image, question, answer) from their test set. As a result, the Vqasynth\_Spacellava and OpenSpaces test sets contain 3,120 and 5,000 image-question-answer pairs, respectively.

Notably, since our model is only trained on the Vqasynth\_Spacellava training set, the OpenSpaces test set can serve as an out-of-distribution benchmark to assess the robustness of spatial reasoning. Furthermore, OpenSpaces is enriched with more diverse and fine-grained spatial VQA questions, whereas Vqasynth\_Spacellava includes a broader mixture of general VQA and spatial reasoning queries. This setup allows us to probe the model’s ability to generalize across both spatial and non-spatial domains.

\subsection{Evaluation Metrics}
\paragraph{Numerical Task.} We adopt three metrics to evaluate the numerical prediction performance: success rate (\%), samples completed (\%), and Symmetric Mean Absolute Percentage Error (sMAPE \%).
Success rate (\%) measures the proportion of predictions that fall within a predefined tolerance range of the ground truth. It reflects how often the model produces ``acceptable'' answers. We define a prediction as success if $\max(\text{GT}/\text{Pred}, \text{Pred}/\text{GT}) < 2$.
Samples Completed (\%) denotes the percentage of test cases where the model outputs a parsable and meaningful numerical value, indicating the reliability and completeness of the model’s responses.
sMAPE (\%) evaluates the prediction accuracy in a scale-invariant manner by comparing predicted values $\hat{y}$ to the ground truth $y$. It is defined as:
\begin{align}
\text{sMAPE} = \frac{1}{n} \sum_{i=1}^{n} \frac{|\hat{y}_i - y_i|}{\left( |\hat{y}_i| + |y_i| \right) / 2 \times 100\%}
\end{align}
To ensure fairness, we only compute sMAPE over test cases where the model produces a valid numerical prediction. Cases without a parseable number are excluded from the average and instead reflected in the ``Samples Completed (\%)'' metric. 
\paragraph{Open-ended Spatial VQA Task.} To evaluate the accuracy of open-ended answers in our Spatial VQA task, we adopt three complementary metrics: BLEU-1~\cite{papineni2002bleu}, Sentence-BERT~\cite{reimers2019sentence} cosine similarity, and an LLM-based evaluation using GPT-4o~\cite{openai2024gpt4o}.
BLEU-1 computes unigram overlaps between the predicted and reference answers. While it captures basic lexical agreement, it often fails to reflect semantic correctness, especially when answers are paraphrased or differ in surface form.
To address this limitation, we additionally report Sentence-BERT similarity, which computes the cosine similarity between sentence-level embeddings of the predicted and reference answers. This embedding-based metric provides a more flexible evaluation of semantic alignment and is better suited for free-form VQA responses where multiple correct phrasings may exist.
Finally, we incorporate an LLM-based evaluation using GPT-4o, which is asked to assess the correctness of the prediction. This metric leverages GPT-4o’s advanced reasoning to provide a human-like judgment.

To better capture the true performance, we categorize questions into types (e.g., binary, descriptive, spatial) and perform type-specific evaluation. This allows us to more accurately assess the model’s behavior under different reasoning demands. To evaluate bounding box predictions in our Spatial VQA task, we adopt two metrics: mean IoU (mIoU) and Accuracy@0.75. mIoU computes the average Intersection over Union between predicted and ground-truth boxes, reflecting overall localization quality. Accuracy@0.75 measures the percentage of predictions with $IoU \geq 0.75$, indicating the proportion of highly accurate localizations. For Yes/No questions, we report standard answer accuracy, measuring the proportion of predictions that exactly match the ground-truth labels. For Distance Estimation, we consider a prediction correct if it falls within the range from 50\% to 200\% of the ground truth, capturing relative correctness under spatial uncertainty.

\begin{table}[t]
  \centering
  \small
  \resizebox{\linewidth}{!}{
  \begin{tabular} {c|c|c|c|c|c|c|c}
    \toprule
     & & \multicolumn{3}{c|}{Metrics} & \multicolumn{3}{c}{ In Range }\\\midrule
   Source & Model & Success Rate(\%)↑ & \makecell{Samples \\Completed(\%)↑ }& sMAPE(\%)↓ &50-100(\%)↑ & 100-150(\%)↑ & 150-200(\%)↑\\ \midrule
    \multirow{2}{*}{Closed} & GPT-4o$^{+}$~\cite{islam2024gpt}& 61.06& - & - & - & - & -\\
    & Gemini-1.5-Flash$^{+}$~\cite{gemini-1.5-flash} & 26.73 & - & - & - & -& -\\ \midrule
    \multirow{6}{*}{Open}& UCSC-VLAA~\cite{chen2025sft} & 25.74 & 99.01& 119.33 & 10.00 & 7.99 & 9.01\\
    & Qwen2.5VL-3B~\cite{bai2025qwen2}$^*$  & 48.51 & 100.00 & 80.93 & \textbf{20.79} & 13.86 & 13.86 \\
    & Qwen2.5VL-3B~\cite{bai2025qwen2} & 21.78 & 92.08 & 117.24 & 9.68 & 5.38 & 8.60 \\
    & \makecell{SpaceThinker-Qwen2.5VL-3B~\cite{SpaceThinker-Qwen2.5VL-3B}} & 27.72 & 98.02 & 125.56 & 16.16 & 7.07 & 5.05 \\
     & InternVL-2.5~\cite{chen2024expanding} & 20.97 & 91.02 & 120.14 & 8.17 & 4.69 & 8.11 \\
    & SVQA-R1 & \textbf{58.42} & \textbf{100.00} & \textbf{68.36} & 18.81 & \textbf{18.81} & \textbf{21.78}\\
    \bottomrule
  \end{tabular}}
  \caption{The test results on Q-Spatial++~\cite{liao2024reasoning}. $*$ indicates models not using the thinking prompt. $+$ means that the results are copied from Q-Spatial++~\cite{liao2024reasoning}. }
\end{table}

\subsection{Evaluation on Numerical Tasks}
Our method, SVQA-R1, outperforms all open-source models across nearly all evaluation metrics and shows a competitive result compared to GPT-4o. In particular, it achieves a 58.42\% success rate, which represents a 31\% absolute improvement over SpaceThinker-Qwen2.5VL-3B, a model fine-tuned with supervised Spatial-CoT data. This demonstrates the effectiveness of our reinforcement learning strategy in enhancing spatial quantitative reasoning.
Moreover, SVQA-R1 surpasses Qwen2.5VL-3B* and Qwen2.5VL-3B by 47\% and 10\% in success rate, respectively. It also achieves the lowest sMAPE (68.36), indicating superior precision in numerical estimation.
These results suggest that reinforcement learning not only improves the model’s ability to align with spatial reasoning tasks but also makes the thinking prompt more adaptable to numerical estimation. Interestingly, we observe that adding the thinking prompt without reinforcement learning (Qwen2.5VL-3B) slightly degrades performance compared to the vanilla Qwen2.5VL-3B$^*$, implying that prompt tuning alone may not benefit spatial reasoning unless coupled with proper optimization signals.

\begin{table}[t]
  \centering
  \small
  \resizebox{\linewidth}{!}{
  \begin{tabular} {c|c|c|c|c|c|c}
    \toprule
    & \multicolumn{3}{c|}{Metrics} & \multicolumn{3}{c}{ In Range }\\\midrule
    Model & Success Rate(\%)↑ & Samples Completed(\%)↑ & sMAP (\%)↓ &50-100(\%)↑ & 100-150(\%)↑ & 150-200(\%)↑\\ \midrule
    Base & 21.78 & 92.08 & 117.24 & 9.68 & 5.38 & 8.60\\
    SFT & 37.62 & 100.00 & 109.13& 16.83 & 6.93 & 13.86\\
    SFT (CoT) & 27.72 & 97.03 & 124.85 & 11.22& 13.27 & 4.08\\ 
    SFT (CoT)$^*$ & 27.72 & 98.02 & 125.56 & 16.16 & 7.07 & 5.05\\ 
    SVQA-R1 & \textbf{58.42} & \textbf{100.00} & \textbf{68.36} & \textbf{18.81} & \textbf{18.81} & \textbf{21.78}\\
    \bottomrule
  \end{tabular}}
  \caption{Compare the SFT and GRPO on Q-Spatial++~\cite{liao2024reasoning}. $*$ indicates the initialization is from UCSC-VLAA~\cite{chen2025sft}, while others are from the Base.}
\end{table}

\begin{table}[t]
  \centering
  \small
  \resizebox{\linewidth}{!}{
  \begin{tabular} {c|c|c|c|c|c|c|c|c|c}
    \toprule
     \multirow{3}{*}{Source}&\multirow{3}{*}{Method} &\multicolumn{4}{c|}{Metrics}&\multicolumn{4}{c}{Spatial-VQA Type}\\\cmidrule{3-10}
     & & \multirow{2}{*}{\makecell{LLM\\ (\%)↑}} & \multirow{2}{*}{\makecell{BLEU-1 \\(\%)↑}}  & \multirow{2}{*}{\makecell{BLEU-2\\ (\%)↑}} & \multirow{2}{*}{sBERT↑} & \multicolumn{2}{c|}{Bbox} & Yes/No & Distance  \\  \cmidrule{7-10} 
     & & & & && mIoU(\%)↑ & Acc.@0.75(\%)↑ & Acc.(\%)↑ & Acc.(\%)↑\\ \midrule
     & & \multicolumn{8}{c}{\textbf{Vqasynth\_Spacellava}} \\\midrule
    \multirow{2}{*}{Closed}    & GPT-4o & 65.92 & 65.02 & 58.50 & 89.23 & 45.02 & 26.50 & 60.05 & 22.51\\
    & Gemini-1.5-Flash~\cite{gemini-1.5-flash} & 31.52 & 34.48 & 27.37 & 63.22 & 34.90 & 20.50 & 49.77 & 14.54\\ \midrule
    \multirow{7}{*}{Open} & QWen2.5VL-3B$^*$ & 45.54  &  33.33 & 31.31 & 68.04 &10.00 & 5.80 & 39.68 & 5.91\\
    & QWen2.5VL-3B & 45.33 & 10.00 & 7.96 & 66.17 & 41.01 &  19.45 & 39.89 & 11.29\\  
    & \text{SpaceThinker{\small-QWen2.5VL-3B}}& 39.39 & 0.00 & 0.00 & 16.14 & 37.00 &  11.60 & 29.37 & 10.75 \\  
    & InternVL-2.5~\cite{chen2024expanding} & 46.69 & 10.30 & 8.20 & 67.49 & 42.24 & 20.23 & 41.09 & 10.95\\
    & UCSC-VLAA~\cite{chen2025sft}& 46.11 & 0.0 & 0.0  & 20.90 & 42.29 & 21.84 &44.21 & 13.44\\
    & SVQA-R1 & \textbf{60.77} & \textbf{63.33} & \textbf{56.43} & \textbf{87.84} & \textbf{43.49} & \textbf{24.57} &  \textbf{58.31} & \textbf{23.66}\\
    \midrule
       & & \multicolumn{8}{c}{\textbf{OpenSpaces}} \\ \midrule
    \multirow{2}{*}{Closed} & GPT-4o & 34.54 &58.70&44.65&89.54&-&-&41.56&15.88 \\
    & Gemini-1.5-Flash~\cite{gemini-1.5-flash} & 20.56 & 45.63 & 30.12 & 75.41 & -&-& 29.10&6.71 \\ \midrule
    
    \multirow{7}{*}{Open}& QWen2.5VL-3B$^*$ & 30.61 &33.33 & 33.33 & 68.05 & -& -& 21.37& 2.18 \\  
    & QWen2.5VL-3B & \textbf{32.49} & 49.71  & 32.35 & \textbf{87.67} & -& - & 29.92 & 3.83\\  
    & SpaceThinker-QWen2.5VL-3B& 16.29 & 1.03 & 1.00 & 31.29 & -& -  & 17.10 & 10.81\\  
    & InternVL-2.5~\cite{chen2024expanding}  & 28.95 & 50.41 & 34.37 & 81.25 & - & - & 21.77 & 4.15 \\
    & UCSC-VLAA~\cite{chen2025sft}& 18.03 &54.68 &37.94 & 87.66 & -& -& 22.16 & 8.15\\
    & SVQA-R1 & 30.97 & \textbf{55.00} & \textbf{41.68} & 85.50 &- & -& \textbf{39.38}& \textbf{14.44} \\
    \bottomrule
  \end{tabular}}
  \caption{Compare the results on the open-ended spatial VQA task on Vqasynth\_Spacellava~\cite{VQASynth} and OpenSpaces~\cite{VQASynth}. $*$ indicates models do not use the thinking prompt.}
  \label{tab:open-end}
\end{table}

\subsection{Evaluation on Open-ended Spatial VQA}
In Figure\ref{tab:open-end}, we compare various methods on the open-ended spatial VQA datasets Vqasynth\_Spacellava and OpenSpaces. Among the open-source models, our proposed SVQA-R1 achieves the best overall performance across nearly all metrics.
Note that the OpenSpaces dataset does not contain bounding box annotations, so the related columns are left blank for that section.


On Vqasynth\_Spacellava, SVQA-R1 shows strong performance, especially on distance estimation. This may benefit from our mirror-consistent training, which encourages the model to maintain consistent spatial interpretation under image flips. Since distance remains invariant under horizontal flipping, this constraint may help the model better learn distance representations. We also observe improvements in bounding box prediction, suggesting that SVQA-R1 enhances spatial relational understanding through view-consistent reasoning.

We observe that the low BLEU-1 scores in SpaceThinker-QWen2.5-VL are primarily caused by a mismatch in answer length and style. Specifically, the model often generates overly brief answers such as ``yes'', ``no'', or ``true'', whereas the ground truth annotations tend to provide slightly more descriptive responses. This discrepancy significantly reduces the unigram precision measured by BLEU-1, even when the model’s answers are semantically correct.

\subsection{Ablation Study}
 From Table \ref{tab:ablation}, we observe that even using only the format reward in GRPO (comparison between (a) and (b)) leads to a significant improvement in performance, especially in Success Rate on Q-Spatial++, which increases by over 12\%. We hypothesize that format guidance helps the model arrive at final answers more reliably by reinforcing output structure during the reasoning process.
On Vqasynth\_Spacellava, while the improvement in LLM accuracy is moderate, BLEU-1 improves dramatically (from 10.00\% to 51.55\%), suggesting that format reward strongly enhances syntactic alignment with the ground truth.
In contrast, using only the semantic reward (comparison between (a) and (c)) improves LLM accuracy (from 45.33\% to 60.34\%) and also yields a large gain in Success Rate on Q-Spatial++. However, BLEU-1 drops to 0.00\% and sBERT to 16.14\%, indicating that the model outputs are semantically close but fail to match token-wise with reference answers. This is expected, as semantic reward emphasizes meaning-level similarity, while BLEU and sBERT are sensitive to surface form and token alignment.
After incorporating spatial-GRPO (e), all metrics improve consistently. The sMAPE is significantly reduced, reflecting better numeric precision in quantity reasoning. Meanwhile, BLEU-1 and sBERT also achieve the highest scores, confirming that SVQA-R1 enhances semantic alignment but also supports more accurate numerical understanding.

 \begin{table}[t]
  \centering
  \small
  \resizebox{\linewidth}{!}{
  \begin{tabular} {l|c|c|c|c|c|c|c|c|c|c}
    \toprule
     & \multicolumn{4}{c|}{Modules} & \multicolumn{3}{c|}{Q-Spatial++}& \multicolumn{3}{c}{Vqasynth\_Spacellava}\\
     \toprule
    Model & Base & \makecell{Format \\ Reward}  & \makecell{Semantic \\ Reward} & \makecell{spatial\\-GRPO} & \makecell{Success \\ Rate (\%) ↑}  & \makecell{Samples \\ Completed (\%) ↑}& sMAPE (\%) ↓ & llm (\%) ↑& BLEU-1 (\%) ↑ & sBERT ↑\\ \midrule
    (a) & \cmark & & & & 21.78 & 92.08& 117.24 & 45.33 & 10.00 & 66.17\\  
    (b) & &\cmark  & & & 33.66 & 94.06 & 107.95 & 49.99 & 51.55 & 85.84 \\  
    (c) & &  &\cmark & & 49.50 & 100.00 & 81.61 & 60.34 & 0.00  & 16.14 \\
    (d) & &\cmark  & \cmark & & 52.48 & 100.00 & 74.49 & 60.35 & 54.29 & 83.51\\
    (e) & &\cmark  & \cmark & \cmark &  \textbf{58.42} & \textbf{100.00} & \textbf{68.36} & \textbf{60.77} & \textbf{63.33} & \textbf{87.84}\\
    \bottomrule
  \end{tabular}}
  \caption{Ablation on reward function and modules on qunatity dataset Q-Spatial++ and open-ended spatial VQA dataset Vqasynth\_Spacellava.}
  \label{tab:ablation}
\end{table}

\subsection{Qualitative Analysis}
Figures~\ref{fig:vis}(a) to (d) are sampled from the Vqasynth\_Spacellava dataset and evaluated using our \textbf{SVQA-R1} model with 3B parameters. Since Vqasynth\_Spacellava is an open-ended spatial VQA dataset, the questions are diverse yet inherently spatial. For example, in Figure~\ref{fig:vis}(b), the model is required to estimate the distance between people distributed across the scene, while in Figure~\ref{fig:vis}(d), it needs to recognize which person is on the ``left'', both requiring spatial relationship reasoning.

From the results, we observe that the model tends to reason step-by-step, exhibiting logical progression rather than generating a flat caption. For instance, it often begins with a clarifying intent such as \textit{``I'll need to ...''}, then introduces uncertainty-aware expressions like \textit{``likely,'' ``would be,''} or \textit{``suggest it could be''} to hypothesize plausible answers. After outlining its reasoning steps, the model often concludes with a summary like \textit{``Therefore, ...''}, indicating a final decision based on prior analysis. Please refer to the supplementary material for more visualization.


\begin{figure*}[tb]
  \centering
   \includegraphics[width=1\linewidth]{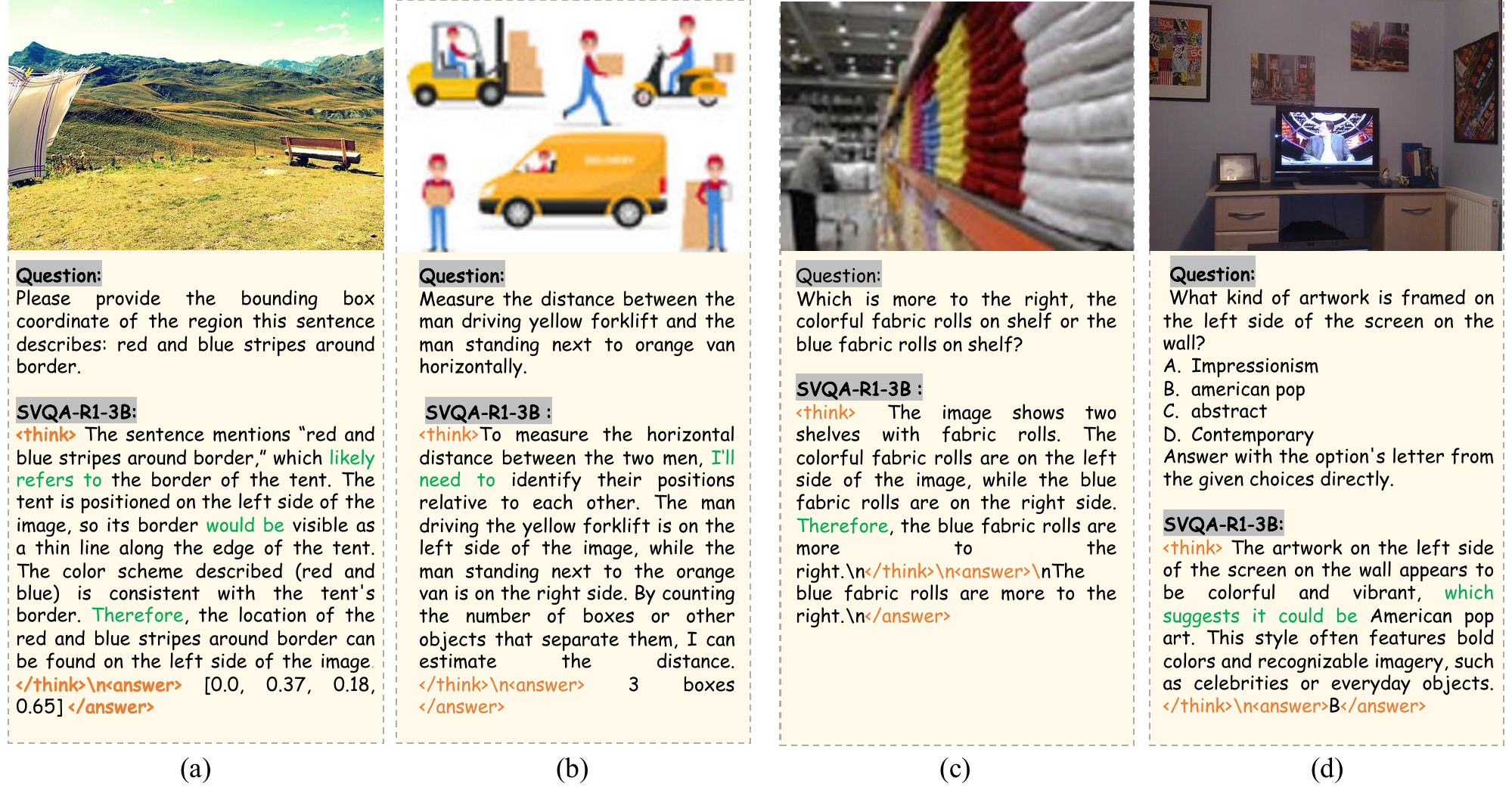}
   \caption{Visualization of different open-ended spatial question-answer types.}
   \label{fig:vis}
\end{figure*}


\section{Conclusion and Limitation}
In this work, we present a novel SVQA-R1 to enhance spatial reasoning capacity in MLLMs. Inspired by the success of the recent R1-style training recipe introduced by Deepseek-R1, we invent a novel view-consistent reward function to encourage the policy model to learn a view-consistent action space. We evaluate our approach on multiple benchmarks, and it exhibits promising results, compared to both closed-source and open-source models. After the self-exploration via Spatial-GRPO in MLLMs, we observe interpretable and reasonable reasoning paths.closed-source and open-source models. 

\textbf{Limitations.} Though the mirror flipping efficiently augments the existing single-view images, more advanced techniques such as Nerf and Gaussian Splatting for novel view synthesis are not investigated yet, where we leave them for future work.


\clearpage
\appendix

\bibliographystyle{abbrv}
\bibliography{neurips_2025}

\begin{thebibliography}{10}

\bibitem{VQASynth}
Spacethinker-qwen2.5vl-3b.
\newblock \url{https://github.com/remyxai/VQASynth}, 2024.

\bibitem{gemini-1.5-flash}
Gemini-1.5-flash.
\newblock \url{https://ai.google.dev/gemini-api/docs/models}, 2025.

\bibitem{SpaceThinker-Qwen2.5VL-3B}
Spacethinker-qwen2.5vl-3b.
\newblock \url{https://huggingface.co/remyxai/SpaceThinker-Qwen2.5VL-3B}, 2025.

\bibitem{bai2025qwen2}
S.~Bai, K.~Chen, X.~Liu, J.~Wang, W.~Ge, S.~Song, K.~Dang, P.~Wang, S.~Wang,
  J.~Tang, et~al.
\newblock Qwen2. 5-vl technical report.
\newblock {\em arXiv preprint arXiv:2502.13923}, 2025.

\bibitem{chen2024spatialvlm}
B.~Chen, Z.~Xu, S.~Kirmani, B.~Ichter, D.~Sadigh, L.~Guibas, and F.~Xia.
\newblock Spatialvlm: Endowing vision-language models with spatial reasoning
  capabilities.
\newblock In {\em Proceedings of the IEEE/CVF Conference on Computer Vision and
  Pattern Recognition}, pages 14455--14465, 2024.

\bibitem{chen2025sft}
H.~Chen, H.~Tu, F.~Wang, H.~Liu, X.~Tang, X.~Du, Y.~Zhou, and C.~Xie.
\newblock Sft or rl? an early investigation into training r1-like reasoning
  large vision-language models.
\newblock {\em arXiv preprint arXiv:2504.11468}, 2025.

\bibitem{chen2024expanding}
Z.~Chen, W.~Wang, Y.~Cao, Y.~Liu, Z.~Gao, E.~Cui, J.~Zhu, S.~Ye, H.~Tian,
  Z.~Liu, et~al.
\newblock Expanding performance boundaries of open-source multimodal models
  with model, data, and test-time scaling.
\newblock {\em arXiv preprint arXiv:2412.05271}, 2024.

\bibitem{daxberger2025mm}
E.~Daxberger, N.~Wenzel, D.~Griffiths, H.~Gang, J.~Lazarow, G.~Kohavi, K.~Kang,
  M.~Eichner, Y.~Yang, A.~Dehghan, et~al.
\newblock Mm-spatial: Exploring 3d spatial understanding in multimodal llms.
\newblock {\em arXiv preprint arXiv:2503.13111}, 2025.

\bibitem{doveh2023dense}
S.~Doveh, A.~Arbelle, S.~Harary, R.~Herzig, D.~Kim, P.~Cascante-Bonilla,
  A.~Alfassy, R.~Panda, R.~Giryes, R.~Feris, et~al.
\newblock Dense and aligned captions (dac) promote compositional reasoning in
  vl models.
\newblock {\em Advances in Neural Information Processing Systems},
  36:76137--76150, 2023.

\bibitem{guo2025deepseek}
D.~Guo, D.~Yang, H.~Zhang, J.~Song, R.~Zhang, R.~Xu, Q.~Zhu, S.~Ma, P.~Wang,
  X.~Bi, et~al.
\newblock Deepseek-r1: Incentivizing reasoning capability in llms via
  reinforcement learning.
\newblock {\em arXiv preprint arXiv:2501.12948}, 2025.

\bibitem{hong2021reinforced}
J.~Hong, P.~Fang, W.~Li, T.~Zhang, C.~Simon, M.~Harandi, and L.~Petersson.
\newblock Reinforced attention for few-shot learning and beyond.
\newblock In {\em Proceedings of the IEEE/CVF Conference on Computer Vision and
  Pattern Recognition}, pages 913--923, 2021.

\bibitem{hudson2019gqa}
D.~A. Hudson and C.~D. Manning.
\newblock Gqa: A new dataset for real-world visual reasoning and compositional
  question answering.
\newblock In {\em Proceedings of the IEEE/CVF conference on computer vision and
  pattern recognition}, pages 6700--6709, 2019.

\bibitem{islam2024gpt}
R.~Islam and O.~M. Moushi.
\newblock Gpt-4o: The cutting-edge advancement in multimodal llm.
\newblock {\em Authorea Preprints}, 2024.

\bibitem{kamath2021mdetr}
A.~Kamath, M.~Singh, Y.~LeCun, G.~Synnaeve, I.~Misra, and N.~Carion.
\newblock Mdetr-modulated detection for end-to-end multi-modal understanding.
\newblock In {\em Proceedings of the IEEE/CVF international conference on
  computer vision}, pages 1780--1790, 2021.

\bibitem{kerbl20233d}
B.~Kerbl, G.~Kopanas, T.~Leimk{\"u}hler, and G.~Drettakis.
\newblock 3d gaussian splatting for real-time radiance field rendering.
\newblock {\em ACM Trans. Graph.}, 42(4):139--1, 2023.

\bibitem{li2024tinyllava}
Y.~Li, Y.~Chen, Y.~Liu, et~al.
\newblock Tinyllava-video-r1: Enhancing video reasoning in small
  vision-language models via reinforcement learning.
\newblock {\em arXiv preprint arXiv:2504.09641}, 2024.

\bibitem{liao2024reasoning}
Y.-H. Liao, R.~Mahmood, S.~Fidler, and D.~Acuna.
\newblock Reasoning paths with reference objects elicit quantitative spatial
  reasoning in large vision-language models.
\newblock {\em arXiv preprint arXiv:2409.09788}, 2024.

\bibitem{liu2023visual}
H.~Liu, C.~Li, Q.~Wu, and Y.~J. Lee.
\newblock Visual instruction tuning.
\newblock {\em Advances in neural information processing systems},
  36:34892--34916, 2023.

\bibitem{ma2024spatialpin}
C.~Ma, K.~Lu, T.-Y. Cheng, N.~Trigoni, and A.~Markham.
\newblock Spatialpin: Enhancing spatial reasoning capabilities of
  vision-language models through prompting and interacting 3d priors.
\newblock {\em Advances in neural information processing systems}, 2024.

\bibitem{mildenhall2021nerf}
B.~Mildenhall, P.~P. Srinivasan, M.~Tancik, J.~T. Barron, R.~Ramamoorthi, and
  R.~Ng.
\newblock Nerf: Representing scenes as neural radiance fields for view
  synthesis.
\newblock {\em Communications of the ACM}, 65(1):99--106, 2021.

\bibitem{openai2024gpt4o}
OpenAI.
\newblock Gpt-4o technical report.
\newblock \url{https://openai.com/index/gpt-4o}, 2024.
\newblock Accessed: 2025-05-14.

\bibitem{papineni2002bleu}
K.~Papineni, S.~Roukos, T.~Ward, and W.-J. Zhu.
\newblock Bleu: a method for automatic evaluation of machine translation.
\newblock In {\em Proceedings of the 40th annual meeting of the Association for
  Computational Linguistics}, pages 311--318, 2002.

\bibitem{patel2024tripletclip}
M.~Patel, N.~S.~A. Kusumba, S.~Cheng, C.~Kim, T.~Gokhale, C.~Baral, et~al.
\newblock Tripletclip: Improving compositional reasoning of clip via synthetic
  vision-language negatives.
\newblock {\em Advances in neural information processing systems},
  37:32731--32760, 2024.

\bibitem{radford2021learning}
A.~Radford, J.~W. Kim, C.~Hallacy, A.~Ramesh, G.~Goh, S.~Agarwal, G.~Sastry,
  A.~Askell, P.~Mishkin, J.~Clark, et~al.
\newblock Learning transferable visual models from natural language
  supervision.
\newblock In {\em International conference on machine learning}, pages
  8748--8763. PmLR, 2021.

\bibitem{reimers2019sentence}
N.~Reimers and I.~Gurevych.
\newblock Sentence-bert: Sentence embeddings using siamese bert-networks.
\newblock In {\em Proceedings of the 2019 Conference on Empirical Methods in
  Natural Language Processing}, pages 3982--3992, 2019.

\bibitem{salter2021attention}
S.~Salter, D.~Rao, M.~Wulfmeier, R.~Hadsell, and I.~Posner.
\newblock Attention-privileged reinforcement learning.
\newblock In {\em Conference on Robot Learning}, pages 394--408. PMLR, 2021.

\bibitem{shao2024deepseekmath}
Z.~Shao, P.~Wang, Q.~Zhu, R.~Xu, J.~Song, X.~Bi, H.~Zhang, M.~Zhang, Y.~Li,
  Y.~Wu, et~al.
\newblock Deepseekmath: Pushing the limits of mathematical reasoning in open
  language models.
\newblock {\em arXiv preprint arXiv:2402.03300}, 2024.

\bibitem{vedantam2019probabilistic}
R.~Vedantam, K.~Desai, S.~Lee, M.~Rohrbach, D.~Batra, and D.~Parikh.
\newblock Probabilistic neural symbolic models for interpretable visual
  question answering.
\newblock In {\em International Conference on Machine Learning}, pages
  6428--6437. PMLR, 2019.

\bibitem{wang2024picture}
J.~Wang, Y.~Ming, Z.~Shi, V.~Vineet, X.~Wang, S.~Li, and N.~Joshi.
\newblock Is a picture worth a thousand words? delving into spatial reasoning
  for vision language models.
\newblock {\em Advances in Neural Information Processing Systems},
  37:75392--75421, 2024.

\bibitem{wang2022raven}
J.~Wang, H.~Zhang, L.~Xie, et~al.
\newblock Raven: Reasoning with visual commonsense for planning in human-robot
  interaction.
\newblock {\em arXiv preprint arXiv:2206.07281}, 2022.

\bibitem{wei2022chain}
J.~Wei, X.~Wang, D.~Schuurmans, M.~Bosma, F.~Xia, E.~Chi, Q.~V. Le, D.~Zhou,
  et~al.
\newblock Chain-of-thought prompting elicits reasoning in large language
  models.
\newblock {\em Advances in neural information processing systems},
  35:24824--24837, 2022.

\bibitem{wu2023reasonrft}
H.~Wu, B.~Zhou, Y.~Wang, et~al.
\newblock Reason-rft: Reasoning-aware reinforcement fine-tuning for
  vision-language models.
\newblock {\em arXiv preprint arXiv:2503.20752}, 2024.

\bibitem{wu2024visualrft}
H.~Wu, B.~Zhou, Y.~Wang, et~al.
\newblock Visual-rft: Visual reinforcement fine-tuning for large
  vision-language models.
\newblock {\em arXiv preprint arXiv:2503.01785}, 2024.

\bibitem{xu2024videor1}
Q.~Xu, Y.~Wang, B.~Zhou, et~al.
\newblock Video-r1: Reinforcing video reasoning in multimodal large language
  models.
\newblock {\em arXiv preprint arXiv:2503.21776}, 2024.

\bibitem{yi2018neural}
K.~Yi, J.~Wu, C.~Gan, A.~Torralba, P.~Kohli, and J.~Tenenbaum.
\newblock Neural-symbolic vqa: Disentangling reasoning from vision and language
  understanding.
\newblock {\em Advances in neural information processing systems}, 31, 2018.

\bibitem{yu2023mvimgnet}
X.~Yu, M.~Xu, Y.~Zhang, H.~Liu, C.~Ye, Y.~Wu, Z.~Yan, C.~Zhu, Z.~Xiong,
  T.~Liang, et~al.
\newblock Mvimgnet: A large-scale dataset of multi-view images.
\newblock In {\em Proceedings of the IEEE/CVF conference on computer vision and
  pattern recognition}, pages 9150--9161, 2023.

\bibitem{zhang2024videochat}
J.~Zhang, Y.~Zhou, Y.~Li, et~al.
\newblock Videochat-r1: Reinforcing spatio-temporal perception in video
  multimodal large language models.
\newblock {\em arXiv preprint arXiv:2504.06958}, 2024.

\bibitem{zhao2023cotvlm}
W.~Zhao, S.~Zheng, H.~Zhang, et~al.
\newblock Cot-vlm: Chain-of-thought prompting for visual language models.
\newblock {\em arXiv preprint arXiv:2309.04761}, 2023.

\bibitem{zhou2024vlm}
B.~Zhou, Y.~Wang, H.~Liu, et~al.
\newblock Vlm-r1: Towards a stable and generalizable r1-style large
  vision-language model.
\newblock {\em arXiv preprint arXiv:2504.07615}, 2024.

\end{thebibliography}

\clearpage
\newpage

\renewcommand{\thesection}{\Alph{section}}
\renewcommand{\thesubsection}{\thesection.\arabic{subsection}}

{\LARGE \textbf{Appendix}}
\setcounter{section}{0} 

\section{Examples of Horizontal Image Flipping}
\subsection{Left-Right Spatial Reasoning}
\begin{figure*}[tbh]
  \centering
   \includegraphics[width=1\linewidth]{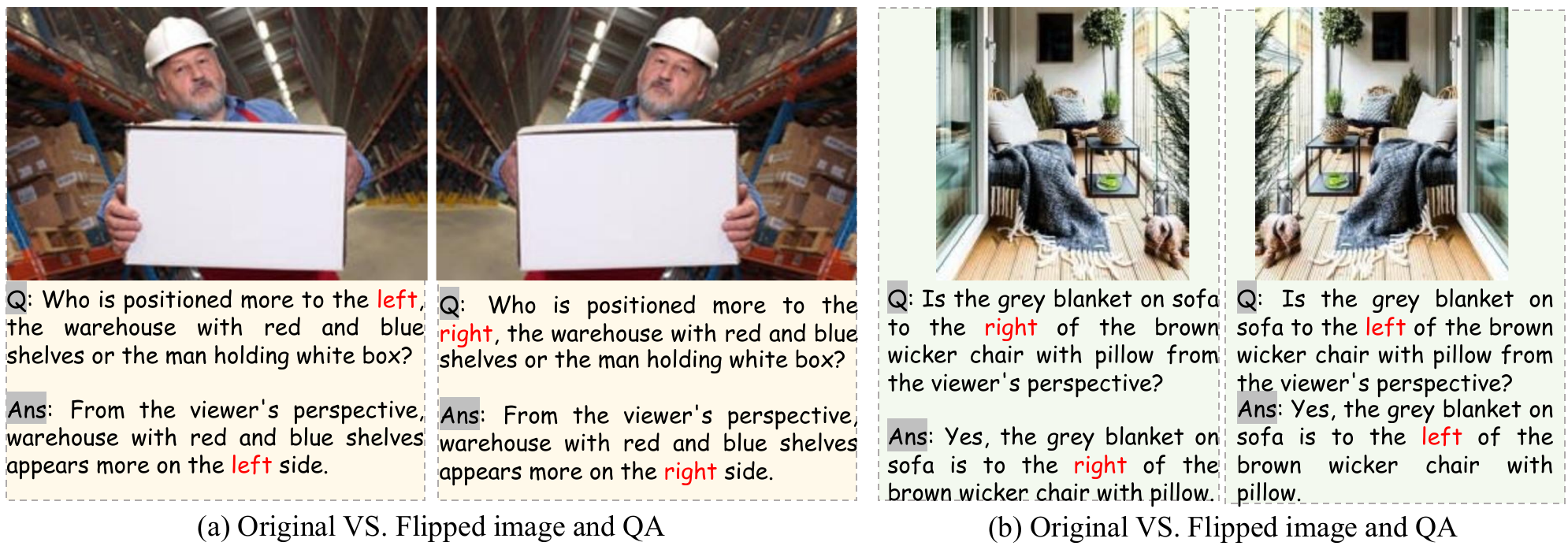}
   \caption{Visualization of original and flipped image and QA for left-right spatial reasoning.}
   \label{fig:vis_1}
\end{figure*}

This category highlights cases where the model's answer depends on reasoning about horizontal spatial relationships, such as determining whether an object is on the left or right. Image flipping directly alters this spatial context, and an accurate model should adapt its response accordingly. These examples in Figure~\ref{fig:vis_1} illustrate the model's ability to reverse its reasoning when the visual orientation changes.

\subsection{Bounding Box Consistency}
\begin{figure*}[tbh]
  \centering
   \includegraphics[width=1\linewidth]{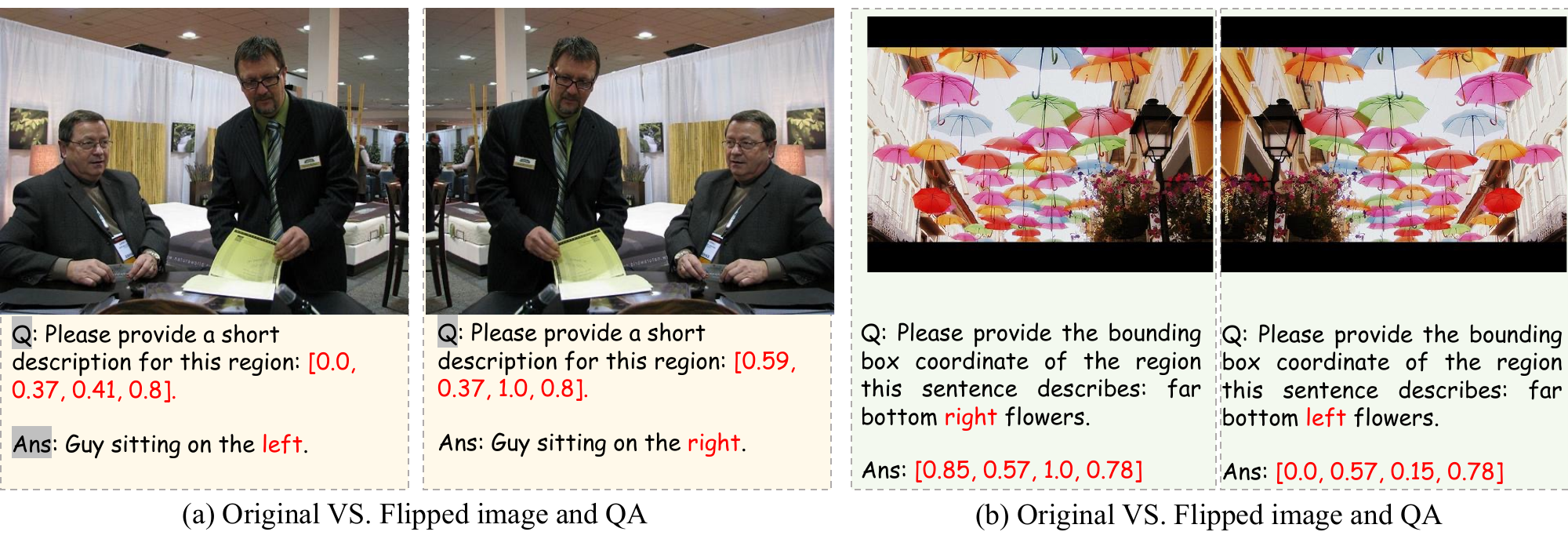}
   \caption{Visualization of original and flipped image and QA  for bounding box consistency.}
   \label{fig:vis_2}
\end{figure*}

In this section, we examine the consistency of predicted bounding boxes under image flipping. A spatially grounded model should correctly adjust the bounding box coordinates to reflect the flipped image layout. We present examples in Figure~\ref{fig:vis_2}, revealing the model’s sensitivity to spatial transformations.

\subsection{Cases Unaffected by Flipping}
\begin{figure*}[tbh]
  \centering
   \includegraphics[width=1\linewidth]{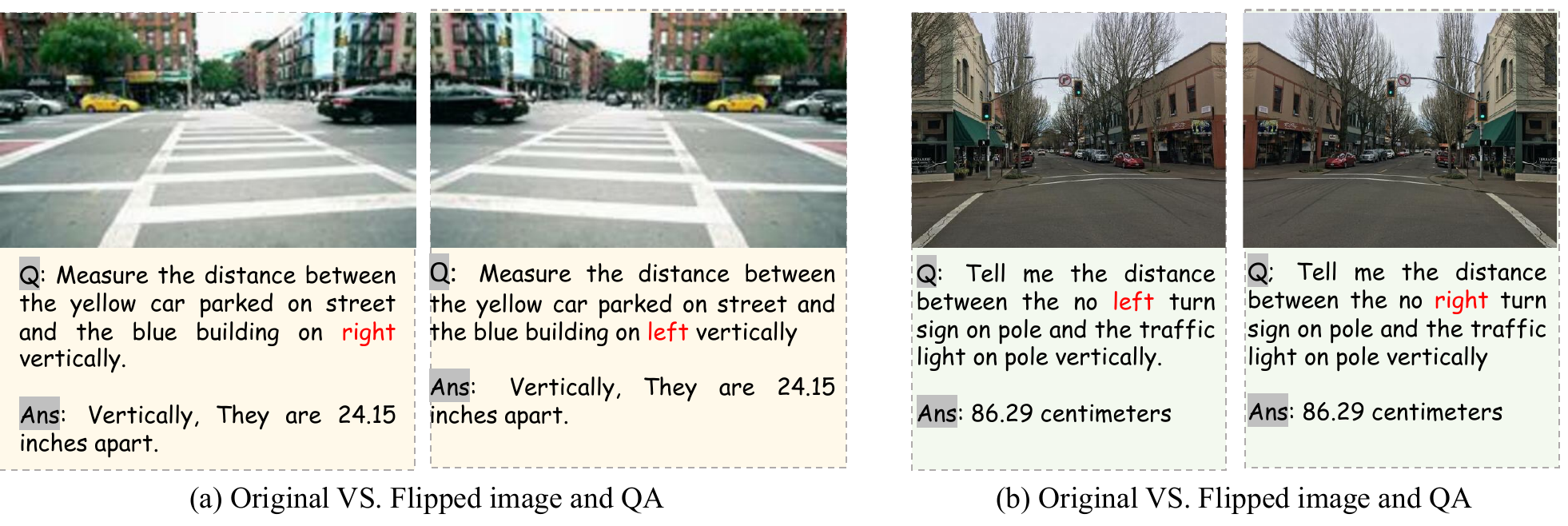}
   \caption{Visualization of original and flipped image and QA for unaffacted cases.}
   \label{fig:vis_3}
\end{figure*}

Some questions are semantically or spatially invariant to horizontal flipping. For example, questions about object attributes, counts, or global scene understanding often yield the same answer regardless of image orientation. This section showcases such cases in Figure~\ref{fig:vis_3}, confirming that the model maintains output stability when the visual change does not affect semantic interpretation.

\section{Determining the Task Type}
In order to evaluate the model's performance more accurately across different types of visual reasoning, we categorize each sample into one of several task types: bounding box prediction, binary (yes/no) classification, or distance estimation. Note that not all examples could be confidently assigned to a specific task type. We exclude ambiguous or unsupported cases from type-specific evaluation. This classification enables us to apply type-specific evaluation metrics and better understand the model's behavior under each reasoning requirement.

In the following subsections, we detail the criteria used to assign a task type to each instance based on its question structure, answer format, or annotated metadata.

 \subsection{Bounding Box}

For bounding box-related questions, we identify them by checking whether the question text contains indicative spatial phrases. Specifically, we use the following keyword list:

\begin{itemize}
  \item \textbf{BBOX\_KEYWORDS:} \texttt{bounding box, box coordinates, coordinates, bbox, where is, x coordinate, y coordinate, draw a box, top left, bottom right, region of.}
\end{itemize}

Using this rule-based classification, we identify \textbf{293 samples} (accounting for \textbf{9.4\%} of the dataset) as bounding box prediction tasks.

 \subsection{Yes/No}
We define alias sets to identify binary (Yes/No) questions and answers, as shown below:

\begin{itemize}
  \item \textbf{YES\_ALIASES:}\texttt{ yes, it is, appears to be, looks like, seems like, definitely, likely, indeed.}
  \item \textbf{NO\_ALIASES:}\texttt{ no, not, doesn't, isn't, unlikely, I don't think, probably not.}
\end{itemize}

Based on these expressions, we classify \textbf{1,742 samples} (accounting for \textbf{34.84\%} of the dataset) as Yes/No questions. Since the model outputs for binary questions are often expressed in natural language rather than strictly as "yes" or "no," we use GPT-4o to assess whether the predicted answer is semantically consistent with the ground truth.

The evaluation is performed via a simple prompt-based classification. For each prediction, we provide GPT-4o with the following instruction, and the \texttt{\{pred\}} and \texttt{\{gt\}} will be replaced by the specific content:

\begin{quote}
\texttt{You are an evaluator. Given a Yes/No question, a ground-truth answer, and a predicted answer, determine whether the predicted answer means the same as the ground-truth answer.\\
Predicted answer: \{pred\}, ground-truth answer: \{gt\}.\\
Output only 0 or 1, where 0 indicates incorrect and 1 indicates correct.}
\end{quote}

 \begin{table}[t]
  \centering
  \small
  \resizebox{\linewidth}{!}{
  \begin{tabular} {c|c| c|c|c|c|c}
    \toprule
      & \multicolumn{3}{c|}{Q-Spatial++}& \multicolumn{3}{c}{Vqasynth\_Spacellava}\\
     \toprule
    $\eta$ &  \makecell{Success \\ Rate (\%) ↑}  & \makecell{Samples \\ Completed (\%) ↑}& sMAPE (\%) ↓ & llm (\%) ↑& BLEU-1 (\%) ↑ & sBERT ↑\\ \midrule
    0 & 52.32 & 100.00 & 73.24 & 59.82 & 53.94 & 81.74\\
    1 &  58.42 & 100.00 &  68.36 & 60.77 & 63.33 & 87.84\\
    2 &  51.49 & 100.00 & 77.66 & 61.47 & 63.34 & 87.86 \\  
    10 &  47.52 & 100.00 & 80.16 & 57.17 & 62.34 & 85.48\\   
    \bottomrule
  \end{tabular}}
  \caption{Performance under different weights applied to the reward difference between original and flipped images.}
  \label{tab:eta}
\end{table}

 \begin{table}[t]
  \centering
  \small
  \resizebox{\linewidth}{!}{
  \begin{tabular} {c|c| c|c|c|c|c}
    \toprule
      & \multicolumn{3}{c|}{Q-Spatial++}& \multicolumn{3}{c}{Vqasynth\_Spacellava}\\
     \toprule
    Model &  \makecell{Success \\ Rate (\%) ↑}  & \makecell{Samples \\ Completed (\%) ↑}& sMAPE (\%) ↓ & llm (\%) ↑& BLEU-1 (\%) ↑ & sBERT ↑\\ \midrule
    Original &  58.42 & 100.00 & 68.36 & 60.77 & 63.33 & 87.84\\   
    VLAA & 53.49 & 100.00 & 76.10 & 61.35 & 56.83 & 83.41 \\   
    \bottomrule
  \end{tabular}}
  \caption{Performance using different base model initializations.}
  \label{tab:init}
\end{table}

 \begin{table}[t]
  \centering
  \small
  \resizebox{\linewidth}{!}{
  \begin{tabular} {l|c|c|c|c|c|c}
    \toprule
      & \multicolumn{3}{c}{Metrics} & \multicolumn{3}{c}{In Range}\\
     \toprule
    Model &  \makecell{Success \\ Rate (\%) ↑}  & \makecell{Samples \\ Completed (\%) ↑}& sMAPE (\%) ↓ &  50-100 (\%) ↑ & 100-150 (\%) ↑& 150-200 (\%) ↑ \\ \midrule
    (a) & 58.42 & 100.00 & 68.36 & 18.81 & 18.81 & 21.78 \\   
    (b) & 20.98 & 100.00 & 184.64 & 5.62 & 7.96 & 10.85   \\   
    \bottomrule
  \end{tabular}}
  \caption{Performance using different prompts for quantity dataset Q-Spatial++.}
  \label{tab:prompt-results}
\end{table}

 \subsection{Distance Prediction}

To identify whether a question requires distance estimation, we rely on a set of indicative phrases commonly associated with spatial measurement. The following keywords are used for matching:

\begin{itemize}
  \item \textbf{DISTANCE\_KEYWORDS:}\texttt{ how far, distance between, distance from, which is closer, which is farther, closer, further, nearer, farthest, measure the distance, what is the distance, spacing between, gap between.}
\end{itemize}

If any of these phrases appear in the question text (matched in a case-insensitive manner), we classify the corresponding sample as a distance prediction task. This classification enables the use of appropriate numerical evaluation metrics, such as absolute error or symmetric Mean Absolute Percentage Error (sMAPE).
Based on this criterion, we identify \textbf{186 samples} (accounting for \textbf{6\%} of the dataset) as distance-related questions. 

\section{The scaling coefficient $\eta$ for the reward difference}
In Table~\ref{tab:eta}, we also explore the effect of applying different scaling coefficients to the reward difference during reward correction. By adjusting the scaling factor of the reward difference, we aim to control the impact of relative feedback and examine its influence on model training and convergence.

We observe that assigning a larger weight to the reward difference improves the performance of the distance estimation task, as reflected by a higher sMAPE score. This indicates that the model produces more accurate numerical predictions. However, we also notice a decrease in the success rate. We hypothesize that while emphasizing the reward gap helps the model focus more on the exact value, it may simultaneously reduce its confidence or stability, leading to uncertainty in determining whether the output is sufficiently close to the ground truth.

In the case of the VQASynth\_Spacellava dataset, we find that increasing the reward weight (comparing weight 1 vs. weight 2) results in slightly better BLEU and Sentence-BERT scores, but a lower LLM-based evaluation score. We conjecture that as the weight increases, the model becomes more sensitive to numerical precision or structured form, but tends to overlook semantic fidelity, which affects alignment with LLM-based judgments.


\section{Different initialization of the MLLMs}
In Table~\ref{tab:init}, we investigate the impact of different base model initializations on downstream performance. By initializing the model from various pretrained checkpoints, we aim to assess how the starting point influences training dynamics and final accuracy. It turns out that the officially original checkpoint performs better on both datasets.

\section{Differnt inference prompt}

We further experiment with different prompting strategies to encourage step-by-step reasoning in the model's responses. Specifically, we compare the original prompt (a) with variants such as spatial reasoning prompts (b) and others tailored to the task context, as summarized in Table~\ref{tab:prompt}, and the results are in Table~\ref{tab:prompt-results}.

We find that although the new prompt encourages the model to produce more reasoning steps, it results in worse performance on the distance estimation task. We hypothesize that this may be due to the fact that the model was originally fine-tuned using the original prompt format, which may not align well with the new prompt structure. This mismatch could lead to degraded performance when adapting to the modified prompting style.

\begin{table}[tbh]
\centering
\begin{tabular}{ >{\centering\arraybackslash}p{2cm} | p{11cm} }
\toprule
\textbf{Prompt Type} & \textbf{Description} \\
\midrule
Original & \{Question\} First output the thinking process in <think> </think> tags and then output the final answer in <answer> </answer> tags.\\
\midrule
Spatial Reasoning Prompt & Question: \{ Question \}
Use the following 4 steps sequentially to answer the question:

Step 1 **Analyze the question**

Step 2 **Identify up to 10 reference scales in the image, ranging from large to small sizes, and list them in the specified format**
- A reference scale must be typical in size.
- A reference scale can be the dimensions of an object or an object part.
- A reference scale must NOT be floor tiles or floor planks.
- Formulate the reference scales using the format: """The [choose from front-to-back, side-to-side, left-to-right, diameter, height (top to bottom edge), or mounting height (bottom edge to floor)] of [object or object part] is approximately [dimension estimate]."""

Step 3 **Propose a robust step-by-step plan to answer the question by using the reference scales in Step 2**
- A robust step-by-step plan performs the estimation in a coarse-to-fine manner.
    - First, use a reliable and large-sized reference scale as the primary reference for estimation.
    - Then, gradually use a reliable and smaller-sized reference scale for adjustment.
    - Repeat until the estimation is precise enough.
- When performing visual comparison, be aware of perspective distortion.
- Do NOT rely on pixel measurements from the images.

Step 4 **Focus on the image and follow the plan in Step 3 to answer the question** \\
\bottomrule
\end{tabular}
\caption{Example of different prompts.}
\label{tab:prompt}
\end{table}

\section{Rescaling Training}

 \begin{table}[h]
  \centering
  \small
  \resizebox{\linewidth}{!}{
  \begin{tabular} {c|c| c|c|c|c|c}
    \toprule
      & \multicolumn{3}{c|}{Q-Spatial++}& \multicolumn{3}{c}{Vqasynth\_Spacellava}\\
     \toprule
    Step &  \makecell{Success \\ Rate (\%) ↑}  & \makecell{Samples \\ Completed (\%) ↑}& sMAPE (\%) ↓ & llm (\%) ↑& BLEU-1 (\%) ↑ & sBERT ↑\\ \midrule
    500 & 43.56 & 100.00 & 84.35 &  60.06 & 52.95  & 84.74\\   
    1000 & 50.50 & 100.00 & 79.32 &  54.25 & 55.36 & 86.51 \\ 
    1500 & 44.55 & 100.00 & 79.31 & 52.13 & 54.29 & 86.23  \\ 
    2000 &  42.57 & 100.00 & 78.57 & 52.37 & 53.38 & 85.64 \\ 
    \bottomrule
  \end{tabular}}
  \caption{Performance in different training steps.}
  \label{tab:step}
\end{table}

We further analyze intermediate checkpoints during training to observe the progression of model performance and reasoning quality. Interestingly, we find that performance initially improves with more training steps but begins to decline after a certain point. This suggests that more training does not always lead to better performance, and there may exist an optimal checkpoint that balances learning and overfitting.

As shown in Table~\ref{tab:step}, we compare selected checkpoints and observe that the model at step 1000 achieves the best performance among them. We believe that evaluating more fine-grained checkpoints could potentially reveal an even better-performing model.

\section{Inferred words in Chain-of-Thought reasoning}
To better understand the nature and quality of the model's reasoning, we conduct a lexical analysis of the generated Chain-of-Thought (CoT) responses. In particular, we examine whether the model produces reasoning-oriented expressions, such as causal connectors, speculative language, and sequential markers, that are commonly associated with multi-step inference.

We define a set of indicative CoT-related keywords spanning various reasoning functions (e.g., uncertainty, causality, step-by-step logic). By counting their occurrences in the generated responses, we aim to quantify the model’s usage of explicit reasoning cues and gain insight into the linguistic characteristics of its CoT outputs.

To quantify the presence of reasoning patterns in model outputs, we define a set of indicative keywords commonly associated with speculative language, causal reasoning, and step-by-step inference:

\begin{itemize}
    \item \textbf{COT\_KEYWORDS:} \texttt{likely, probably, possibly, maybe, might be, could be, seems, appears to, I think, I guess, I'm not sure, because, since, therefore, thus, so, hence, as a result, that means, which implies, accordingly, first, next, then, finally, in the first step, in the second step, after that, subsequently, clearly, obviously, evidently, definitely, in fact, it is important to note, if, suppose, assuming that, in case, let's say, consider that.}
\end{itemize}

Based on this keyword list, we observe that the average number of reasoning-related words per response is \textbf{0.8}, indicating that most responses contain at least one token suggestive of inferred or stepwise reasoning.

\section{More Visualization Examples}
\begin{figure*}[tbh]
  \centering
   \includegraphics[width=1\linewidth]{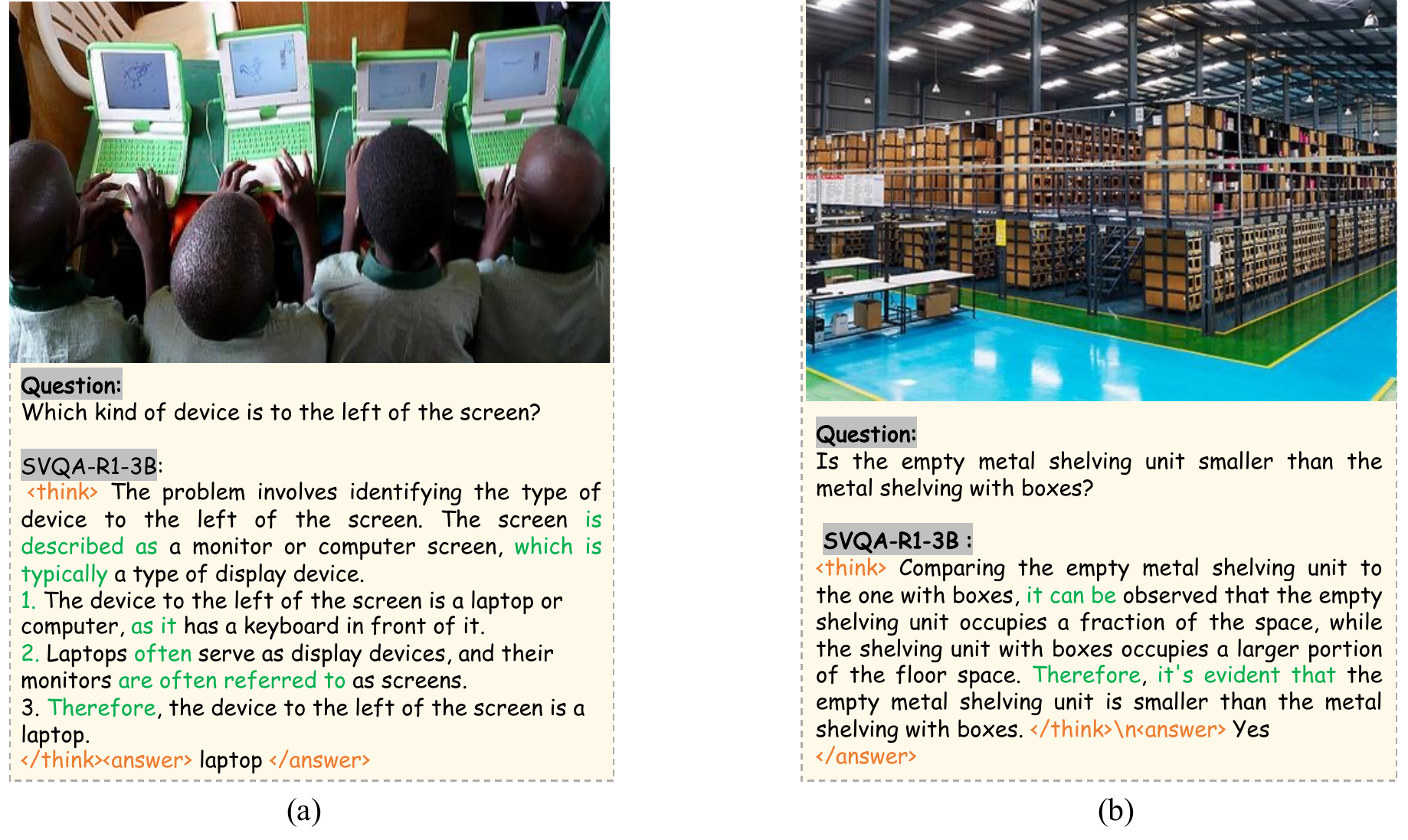}
   \label{fig:vis}
\end{figure*}

\begin{figure*}[h]
  \centering
   \includegraphics[width=1\linewidth]{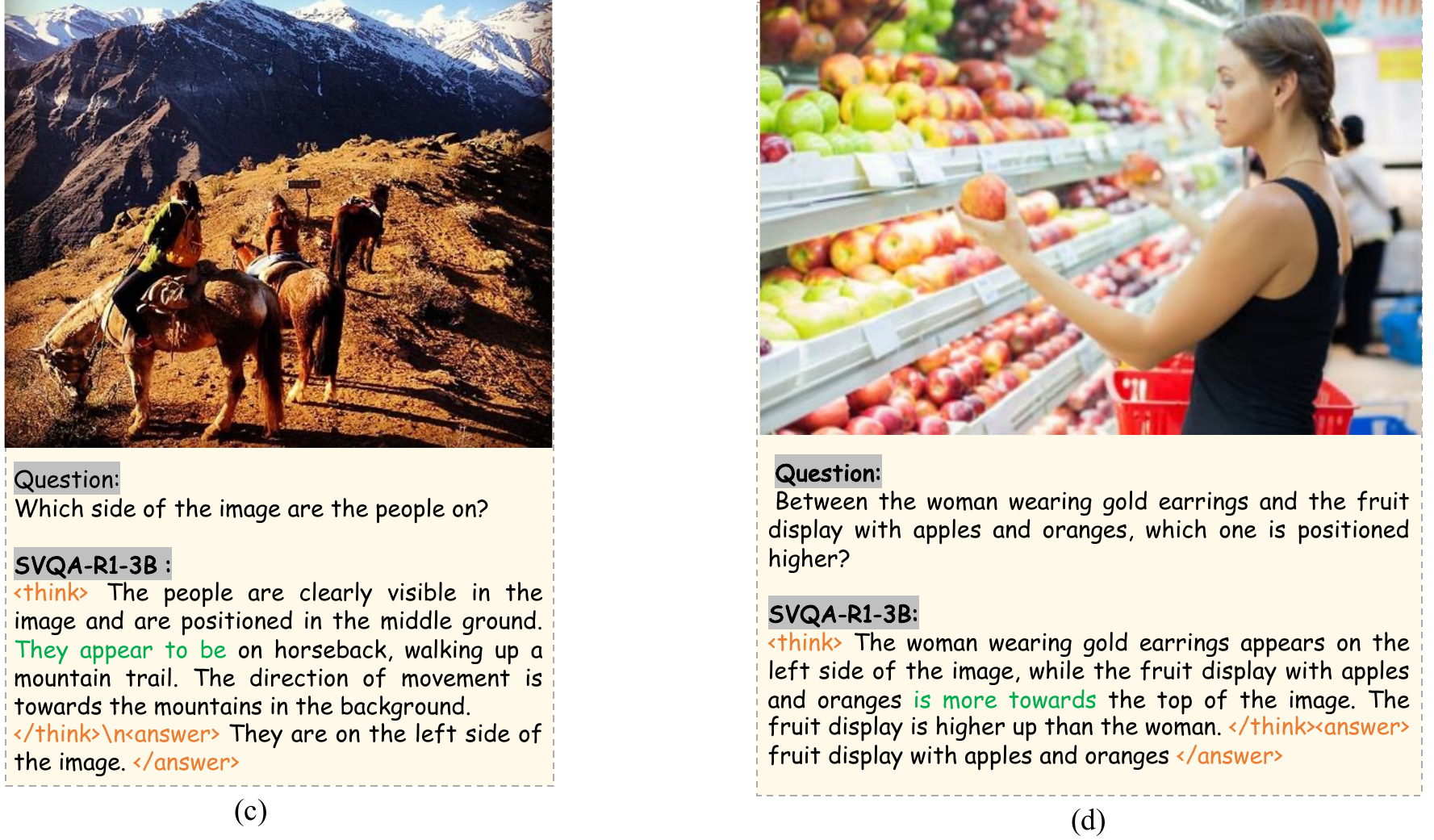}
   \label{fig:vis}
\end{figure*}

\begin{figure*}[h]
  \centering
   \includegraphics[width=1\linewidth]{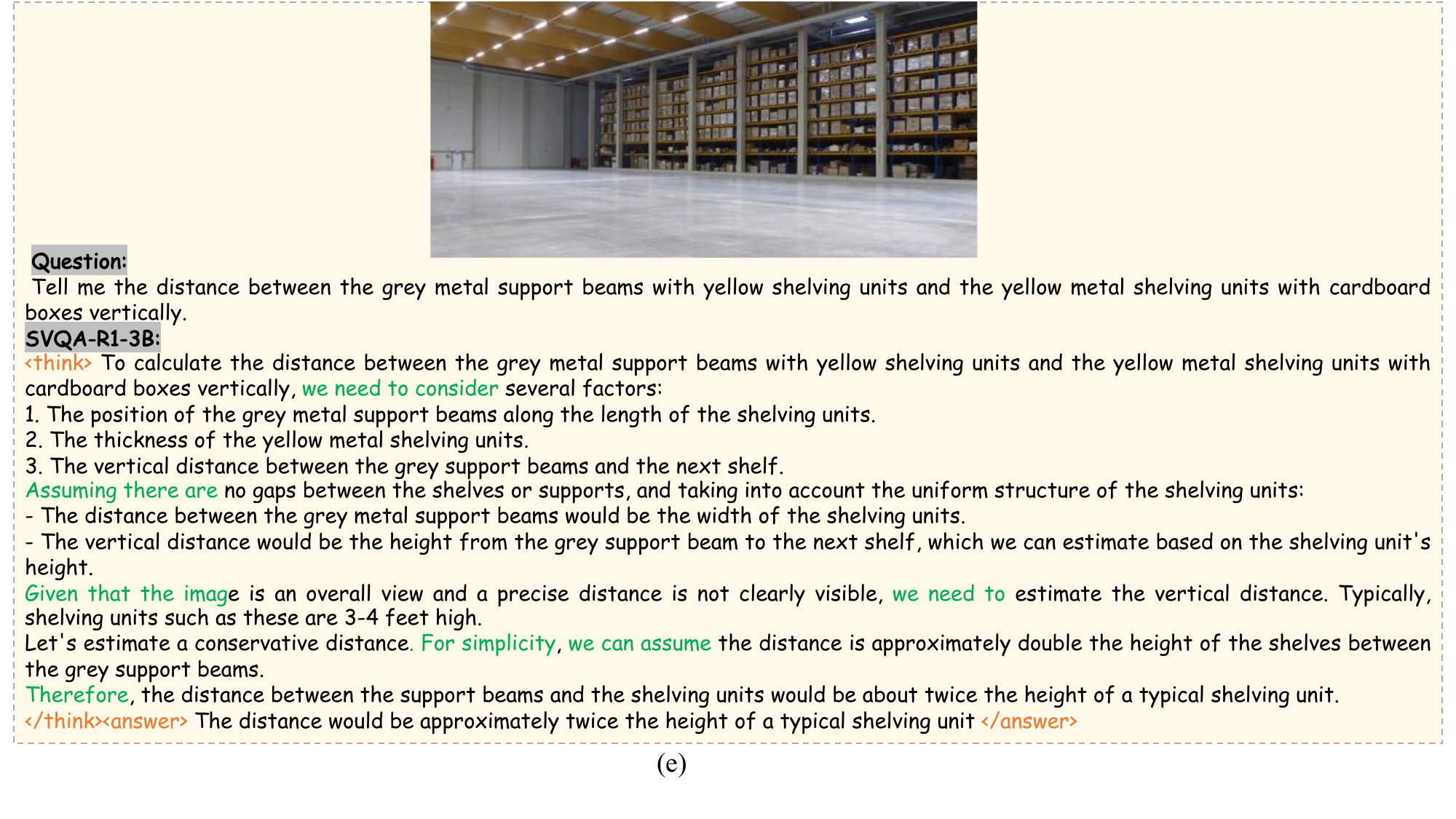}
   \caption{More visualization examples with thinking evidence.}
   \label{fig:vis}
\end{figure*}

\end{document}